\documentclass{article}

\PassOptionsToPackage{square, numbers}{natbib}
\usepackage[final]{neurips_2023}

\usepackage[utf8]{inputenc} % allow utf-8 input
\usepackage[T1]{fontenc}    % use 8-bit T1 fonts
\usepackage{hyperref}       % hyperlinks
\usepackage{url}            % simple URL typesetting
\usepackage{booktabs}       % professional-quality tables
\usepackage{amsfonts}       % blackboard math symbols
\usepackage{nicefrac}       % compact symbols for 1/2, etc.
\usepackage{microtype}      % microtypography

\usepackage{wrapfig}
\usepackage{graphicx}
\usepackage{amsmath}
\usepackage{amssymb}
\usepackage{booktabs}
\usepackage{bm}
\usepackage{caption}
\usepackage{float}
\usepackage{blindtext}
\usepackage{array}
\usepackage{tabularx}
\usepackage{booktabs}
\usepackage{ragged2e}
\usepackage{color, colortbl}
\usepackage{makecell}
\usepackage{caption}
\usepackage{bbding}

\definecolor{Gray}{gray}{0.9}
\definecolor{LightCyan}{rgb}{0.88,1,1}

\newcolumntype{P}[1]{>{\centering\arraybackslash}p{#1}}
\newcolumntype{M}[1]{>{\centering\arraybackslash}m{#1}}
\newcommand{\Mat}{\boldsymbol}

\newcommand{\real}{\mathbb{R}}

\newcommand\mypara[1]{\vspace{0mm}\noindent\textbf{#1}}
\usepackage{multirow}
\usepackage[normalem]{ulem}
\useunder{\uline}{\ul}{}
\usepackage[capitalize]{cleveref}
\crefname{section}{Sec.}{Secs.}
\Crefname{section}{Section}{Sections}
\Crefname{table}{Table}{Tables}
\crefname{table}{Tab.}{Tabs.}

\newcommand*\samethanks[1][\value{footnote}]{\footnotemark[#1]}

\title{ReTR: Modeling Rendering Via Transformer for Generalizable Neural Surface Reconstruction}

\author{Yixun Liang\textsuperscript{$1$}\thanks{Equal contribution.} \quad
        Hao He\textsuperscript{$1,2$}\samethanks \quad
        Ying-Cong Chen\textsuperscript{$1,2$}\thanks{Corresponding author} \quad \\
            \small$^1$ The Hong Kong University of Science and Technology~(Guangzhou).\quad \\ 
            \small$^2$ The Hong Kong University of Science and Technology.\\
            \small\texttt{yliang982@connect.hkust-gz.edu.cn, hheat@connect.ust.hk, yingcongchen@ust.hk} \\ 
}

\begin{document}

\maketitle

\begin{abstract}
Generalizable neural surface reconstruction techniques have attracted great attention in recent years. However, they encounter limitations of low confidence depth distribution and inaccurate surface reasoning due to the oversimplified volume rendering process employed. In this paper, we present Reconstruction TRansformer (ReTR), a novel framework that leverages the transformer architecture to redesign the rendering process, enabling complex render interaction modeling. It introduces a learnable \textit{meta-ray token} and utilizes the cross-attention mechanism to simulate the interaction of rendering process with sampled points and render the observed color. Meanwhile, by operating within a high-dimensional feature space rather than the color space, ReTR mitigates sensitivity to projected colors in source views. Such improvements result in accurate surface assessment with high confidence. We demonstrate the effectiveness of our approach on various datasets, showcasing how our method outperforms the current state-of-the-art approaches in terms of reconstruction quality and generalization ability.  \textit{Our code is available at} \url{https://github.com/YixunLiang/ReTR}.

\end{abstract}

\section{Introduction}
In the realm of computer vision and graphics, extracting geometric information from multi-view images poses a significant challenge with far-reaching implications for various fields, including robotics, augmented reality, and virtual reality. As a popular approach to this problem, neural implicit reconstruction techniques~\cite{yariv2021volume,wang2021neus,oechsle2021unisurf,niemeyer2020differentiable,darmon2022improving,wang2022hf} are frequently employed, generating accurate and plausible geometry from multi-view images by utilizing volume rendering and neural implicit representations based on the Sign Distance Function~(SDF)~\cite{park2019deepsdf} and its variant. Despite their efficacy, these methods possess inherent limitations such as the lack of cross-scene generalization capabilities and the necessity for extensive computational resources for training them from scratch for each scene. Furthermore, these techniques heavily rely on a large number of input views.

\begin{figure*}[t]
%\vspace{-2em}
%\vskip 0.2in
\begin{center}
\setlength{\abovecaptionskip}{0.cm}
\centerline{\includegraphics[width=\textwidth]{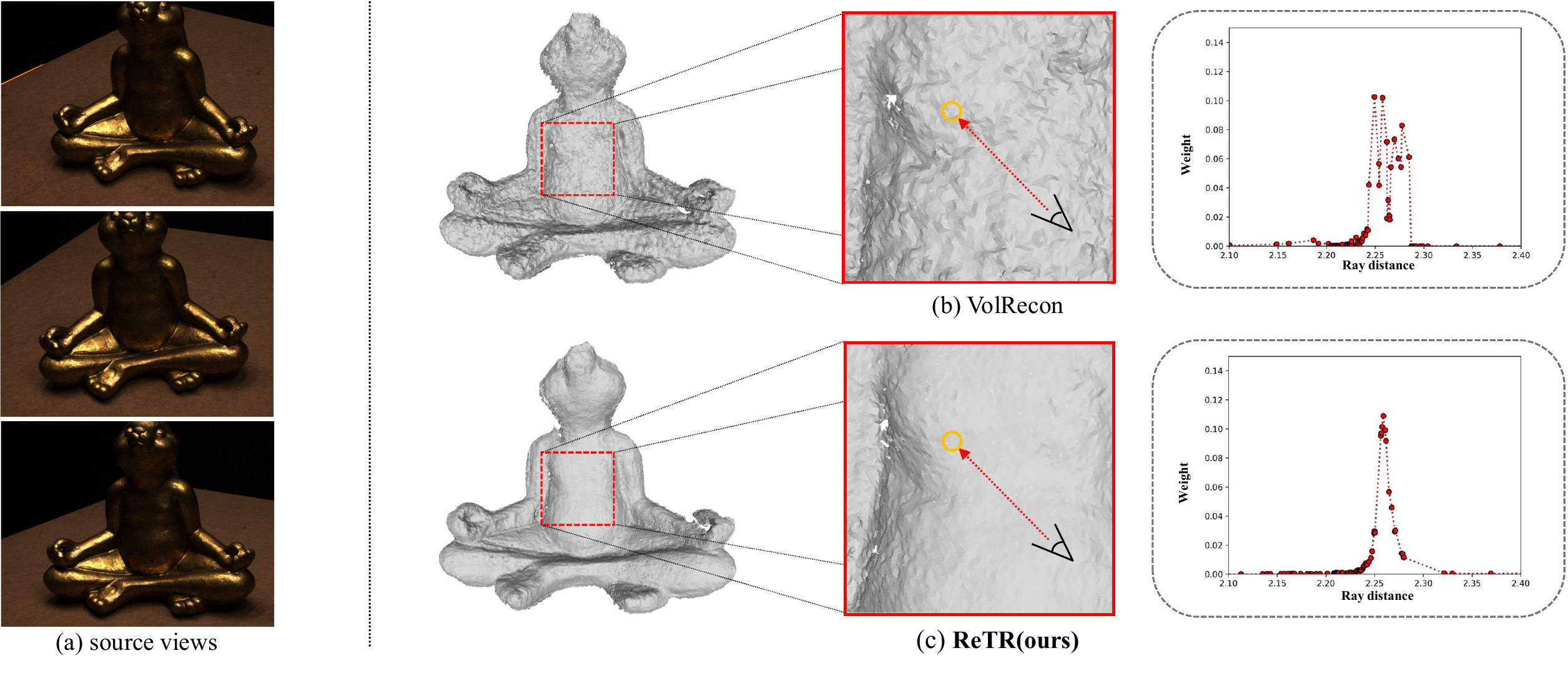}}
\caption{Generalizable neural surface reconstructions from three input views in~(a). VolRecon~\cite{ren2022volrecon} produces depth distribution with low kurtosis and a noisy surface as shown in~(b). In contrast, our proposed \textbf{ReTR} successfully extracts plausible surfaces with sharper depth distribution that has high kurtosis as shown in~(c).}
\label{fig:teaser}
\vspace{-1.1cm}
\end{center}
\end{figure*}
Recent studies such as SparseNeuS~\cite{long2022sparseneus} and VolRecon~\cite{ren2022volrecon} have attempted to overcome these challenges by integrating prior image information with volume rendering methods, thereby achieving impressive cross-scene generalization capabilities while only requiring sparse views as input. Nevertheless, these solutions are still based on volume rendering, which poses some intrinsic drawbacks in surface reconstruction. Specifically,  volume rendering is a simplification of the physical world and might not capture the full extent of its complexity. It models the interactions of incident photons and particles into density, which is predicted solely based on sampled point features. This oversimplified modeling fails to disentangle the contribution of the light transport effect and surface properties to the observed color, resulting in an inaccurate assessment of the actual surface. Moreover, the prediction of color blending heavily relies on the projected color of the source views, thereby overlooking intricate physical effects. These shortcomings can lead to less confident surface predictions, producing a depth distribution with low kurtosis, and may be accompanied by high levels of noise, thereby compromising the overall quality of the reconstruction as illustrated in Fig.~\ref{fig:teaser}~(b). 

In this paper, we first propose a more generalized formulation for volume rendering. Then based on this formulation, we introduce the Reconstruction TRansformer (ReTR), a novel approach for generalizable neural surface reconstruction. Our method utilizes the transformer to redesign the rendering process, allowing for accurate modeling of the complex render interaction while retaining the fundamental properties that make volume rendering effective. Particularly, we propose a learnable token, the \textit{meta-ray token}, that encapsulates the complex light transport effect. By leveraging this token and the "cross-attention" mechanism, we simulate the interaction of rendering with the sampled points and aggregate the features of each point to render the observed color in an end-to-end manner. Additionally, we introduce a unidirectional transformer and continuous positional encoding to simulate photon-medium interaction, effectively considering occlusion and the interval of sample points. Moreover, as our model operates in the feature space rather than the color space, to further enhance the accuracy of semantic information and enable seamless adaptation, we propose a novel hybrid extractor designed to efficiently extract 3D-aware features.

Our method provides an excellent solution for generalizable neural surface reconstruction, offering several advantages over traditional volume rendering techniques. First, the ability to generalize complicated physical effects in a data-driven way provides an efficient approach to decoupling the contribution of light transport and surface to the observed color. Second, the entire process takes place in a high-dimensional feature space, rather than the color space, reducing the model's sensitivity to projected color in source views. Moreover, the transformer employs a re-weighted \textit{softmax} function of the attention map, driving the learning of a depth distribution with positive kurtosis. As a result, our method achieves a more confident distribution, leading to reduced noise and improved quality, As shown in Fig.~\ref{fig:teaser}~(c).

In summary, our contribution can be summarized as:
\begin{itemize}
    \item We identify the limitation and derive a general form of volume rendering. By leveraging this form, we can effectively tailor the rendering process for task-specific requirements.
    \item Through the derived general form, we propose ReTR, a learning-based rendering framework utilizing transformer architecture to model light transport. ReTR incorporates continuous positional encoding and leverages the hybrid feature extractor to enhance performance in generalizable neural surface reconstruction.
    \item Extensive experiments conducted on DTU, BlendedMVS, ETH3D, and Tanks \& Temples datasets~\cite{aanaes2016large, yao2020blendedmvs, Knapitsch2017,schops2017multi} validate the efficacy and generalization ability of ReTR.
\end{itemize}

% %%%%%%%%%%%%%%%%%   Related Work    %%%%%%%%%%%%%%%%%%%%%%%%%%
\section{Related Works}
\label{sec:related}
\paragraph{Multi-View Stereo (MVS).} Multi-view stereo methods is another branch for 3D reconstruction, which can be broadly categorized into three main branches: depth maps-based~\cite{campbell2008using, chang2022rc, schonberger2016pixelwise,gu2020cascade,yang2020cost, luo2019p}, voxel grids-based~\cite{kutulakos2000theory, seitz1999photorealistic, yao2019recurrent}, and point clouds-based~\cite{furukawa2009accurate, lhuillier2005quasi, chen2019point}. Among these, depth maps-based methods are more flexible and hence more popular than the others. Depth maps-based methods typically decouple the problem into depth estimation and fusion, which has shown impressive performance with densely captured images. However, these methods exhibit limited robustness in situations with a shortage of images, thereby highlighting the problem we aim to address.

\mypara{Neural Surface Reconstruction.} With the advent of NeRF~\cite{mildenhall2021nerf}, there has been a paradigm shift towards using similar techniques for shape modeling, novel view synthesis, and multi-view 3D reconstruction. IDR~\cite{yariv2020multiview} uses surface rendering to learn geometry from multi-view images, but it requires extra object masks. Several methods~\cite{wang2021neus, oechsle2021unisurf, yariv2021volume,zhang2021learning, ge2023ref} have attempted to rewrite the density function in NeRF using SDF and its variants, successfully regressed plausible geometry. Among them, NeuS~\cite{wang2021neus} first uses SDF value to model density in the volume rendering to learn neural implicit surface, offering a robust method for multi-view 3D reconstruction from 2D images. However, these methods require lengthy optimization to train each scene independently. 
Inspired by the recent success of generalizable novel view synthesis~\cite{wang2021ibrnet, chen2021mvsnerf, yu2021pixelnerf, trevithick2021grf,peng2020convolutional}. SparseNeuS~\cite{long2022sparseneus} and VolRecon~\cite{ren2022volrecon} achieve generalizable neural surface reconstruction using the information from the source images as the prior to neural surface reconstruction. However, these methods suffer from oversimplified modeling of light transport in traditional volume rendering and color blending, resulting in extracted geometries that fail to make a confident prediction of the surface, leading to distortion in the reconstructions. However, there are also some attempts~\cite{zhang2021physg,zhang2021nerfactor,knodt2021neural} that adopt other rendering methods in implicit neural representation, such as ray tracing. However, these models are still based on enhanced explicit formulations; thus, their capacity to model real-world interaction is still limited. In contrast, our method introduces a learning-based rendering, providing an efficient way to overcome such limitations. 

\mypara{Learning Based Rendering.} Unlike volume rendering, another line of work~\cite{sitzmann2019deepvoxels, sitzmann2021light, sajjadi2022scene} explores deep learning techniques to simulate the rendering process. Especially Recurrent neural networks, which naturally fit the rendering process. Specifically, DeepVoxels~\cite{sitzmann2019deepvoxels} employs GRU to process voxel features along a ray, and its successor SRN~\cite{sitzmann2019scene}, leverages LSTM for ray-marching. However, such methods recursively process features and demand large computation resources. The computation constraints inhibit such methods' ability to produce high-quality renderings. Unlike RNN-based methods, ReTR leverages the transformer to parallel compute each point's hitting probability. Greatly improve the efficiency of the rendering process and achieve high-quality renderings.

\mypara{Transformers With Radiance Field.} The attention mechanism in transformers~\cite{vaswani2017attention} has also been widely used in the area of radiance field. In image-based rendering, IBRNet~\cite{wang2021ibrnet} proposes a ray transformer to process sampled point features and predict density. NeRFormer~\cite{reizenstein2021common} utilizes a transformer to aggregate source views and construct feature volumes. NeuRays~\cite{liu2022neuray} leverages neural networks to model and address occlusions, enhancing the quality and accuracy of image-based rendering. GPBR~\cite{suhail2022generalizable} employs neural networks to transform and composite patches from source images, enabling versatile and realistic image synthesis across various scenes. However, these methods only use the transformer to enhance feature aggregation, and the sampled point features are still decoded into colors and densities and aggregated using traditional volume rendering, leading to unconfident surface prediction. Recently, GNT~\cite{wang2022attention} naively replaces classical volume rendering with transformers in image-based rendering, which overlooks the absence of occlusion and positional awareness within the transformer architecture. In contrast to GNT, we improved the traditional transformer architecture in those two limitations based on an in-depth analysis of the fundamental components to make volume rendering work. 

%%%%%%%%%%%%%%%%%   Method    %%%%%%%%%%%%%%%%%%%%%%%%%%
\section{Methodology}
In this section, we present an analysis of the limitations of existing generalizable neural surface reconstruction approaches that adopt volume rendering from NeRF~\cite{mildenhall2021nerf} and propose ReTR, a novel architecture that leverages transformer to achieve learning-based rendering. We introduce the formulations of volume rendering and revisit its limitations in generalizable neural surface reconstruction in Sec~\ref{sec:problem_definition}. We then depict the general form of volume rendering in Sec.~\ref{sec:general_form} and present our learning-based rendering in Sec.~\ref{sec:trans_deocde}. To effectively extract features for our proposed rendering, we further introduce the hybrid extractor in Sec.~\ref{sec.hbe}. Our loss functions are explained in Sec.~\ref{sec:netarch}.

\label{sec:general_form}
\begin{figure*}[t]
%\vskip 0.2in
%\vspace{-2em}
\begin{center}
\setlength{\abovecaptionskip}{0.cm}
\centerline{\includegraphics[width=1.\textwidth]{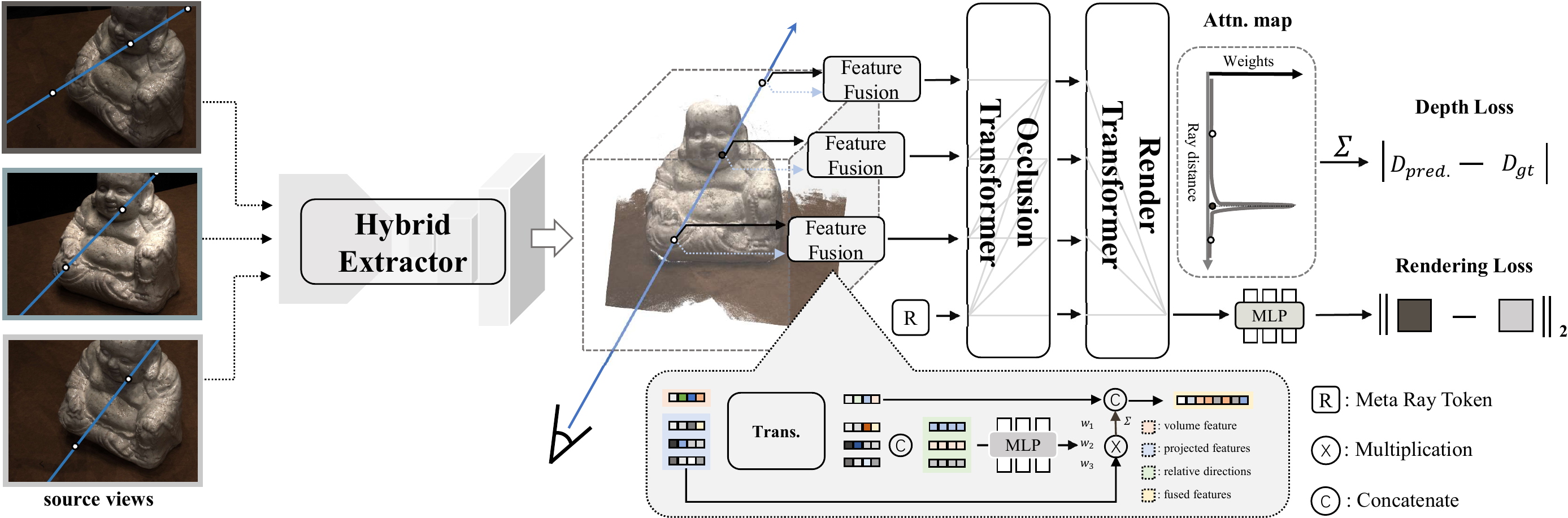}}
\vspace{0.2cm}
\caption{Our ReTR pipeline comprises several steps: (1). Extracting features through the proposed hybrid extractor model from source views, (2). Processing features in each sample point using the feature fusion block, and (3). Using the occlusion transformer and render transformer to aggregate features along the ray and predict colors and depths.}
\label{fig:overview}
\vspace{-0.5cm}
\end{center}
\end{figure*}
\subsection{Preliminary}
\label{sec:problem_definition}
Generalizable neural surface reconstruction aims to recover the geometry of a scene $\Mat{S}$, which is represented by a set of $M$ posed input views~$\Mat{S} = \{\Mat{I}_j,\Mat{P}_j\}_{j=1}^{M}$, where $\Mat{I}_j \in \real^{H \times W \times 3}$ and $\Mat{P}_j \in \real^{3 \times 4}$ are the $j$-th view's image and camera parameters, respectively.
% Generalizable multi-view reconstruction aims to recover the geometric of scene $S$ from the prior information of given posed views $\left\{\Mat{I}_1, \cdots, \Mat{I}_M\right\}$.
Existing approaches~\cite{long2022sparseneus,ren2022volrecon} generate the features of the radiance field from $\Mat{S}$ using a neural network~$\mathcal{F}_{enc.}$, and we formalize the process as:
\begin{equation}
\label{eq:1}
\mathbf{f}^{v},\{\mathbf{f}^{img}_1,\dots,\mathbf{f}^{img}_M\} = \mathcal{F}_{enc.}(\Mat{S}),
\end{equation}
where ~$\mathbf{f}^{v} \in \real^{R\times R \times R \times D}$ is the volume feature with resolution $R$ and ~$\mathbf{f}_{j}^{img}\in \real^{h\times w\times D}$ is the image feature with dimension $D$.
To decode the color of a sampled ray $\mathbf{r} = (\mathbf{o},\mathbf{d})$ passing through the scene, where $\mathbf{o}$ and $\mathbf{d}$ denotes as ray original and direction, the existing approaches~\cite{long2022sparseneus,ren2022volrecon} sample $N$ points along the ray from coarse to fine sampling between near and far planes and obtain the location $\Mat{x} \in \real^3$ of each sample point:
\begin{equation}
\label{eq:2}
\Mat{x}_i = \mathbf{o}+t_i\mathbf{d}, \quad i = 1, \dots, N.
\end{equation}
The predicted SDF from features first converts to weights using the conversion function~\cite{wang2021neus}, denoted as $\sigma_{i}$. The weights are then used to accumulate the re-weighted projected colors along the ray using volume rendering. Specifically:
\begin{equation}
\label{eq:volume_rendering}
C(\mathbf{r})=\sum_{i=1}^{N} T_{i}\left(1-\exp \left(-\sigma_{i}\right)\right) \mathbf{c}_{i}, \quad \text{where} \quad T_{i}=\exp \left(-\sum_{j=1}^{i-1} \sigma_{j}\right),
\end{equation}
\begin{equation}
\label{eq:weighted_color}
\mathbf{c}_{i} = \sum_{j=1}^{M}\mathcal{F}_{weight.}(\mathbf{f}^v_i,\{\Pi(\mathbf{f}^{img}_k,\Mat{x}_i)\}_{k=1}^{M})\Pi(\Mat{I}_j,\Mat{x}_i).
\end{equation}
Here the $\mathcal{F}_{weight.}(\cdot)$ denotes the module that predicts the weight of each projected color. The $\mathbf{f}^v_i$ represents the volume feature at $\Mat{x}_i$ obtained using trilinear interpolation in $\mathbf{f}^v$. $\Pi(\Mat{I},\Mat{x})$ denotes the operation that projects  $\Mat{x}$ onto the corresponding input grid $\Mat{I}$ and then extracts the feature at the projected location through bilinear interpolation.

\mypara{Limitation.} The volume rendering, denoted by Eq.~\eqref{eq:volume_rendering}, greatly simplifies light transport processes, thus introducing key limitations. A complete physical model for light transport categorizes the process into absorption, emission, out-scattering, and in-scattering. Each represents a nonlinear photon-particle interaction, with incident photon properties being influenced by both their inherent nature and medium characteristics. However, Eq.~\eqref{eq:volume_rendering} condenses these complexities into a single density value, predicted merely based on sampled point features, leading to an \textit{oversimplification of incident photon modeling}. Moreover, the color determination method, influenced by a weighted blend of projected colors akin to Eq.~\eqref{eq:weighted_color}, \textit{over-relies on input view projected colors, overlooking intricate physical effect}s. As a result, the model requires a wider accumulation of projected colors from points near the exact surface, resulting in a "murky" surface appearance, as depicted in Fig.~\ref{fig:teaser}.

% \subsection{Transformer as a Differentiable Renderer}
\subsection{Generalized Rendering Function}
To overcome the limitations we mentioned in Sec.~\ref{sec:problem_definition}, we propose an improved rendering equation that takes into consideration of incident photon modeling and feature-based color decoding. As we revisit Eq.~\eqref{eq:volume_rendering} and determine its key components to facilitate our redesign. This differentiable equation consists of three parts: The $T_{i}$ term accounts for the accumulated transmittance and gives the first surface a bias to contribute more to the observed color. The $\left(1-\exp \left(-\sigma_{i}\right)\right)$ term denotes the alpha value of traditional alpha compositing, which represents the transparency of $\Mat{x}_i$ and is constrained to be non-negative. The color part of Eq.~\eqref{eq:volume_rendering} denotes the color of $\Mat{x}_i$. Based on these analyses, we summarize three key rendering properties that the system must hold: 
\begin{enumerate}
\item \textit{Differentiable}. To enable effective learning, the weight function needs to be differentiable to the training network with observed color through back-propagation. 
\item \textit{Occlusion-aware}. In line with the bias towards the first surface in the original equation, the weight function needs to be aware that the points close to the first \textit{exact surface} should have a larger contribution to the final output color than other points.
\item \textit{Non-negative}. Echoing the non-negativity constraint in the alpha value, the weight of each point also needs to be positive.
\end{enumerate}

Having identified these key properties, we can now reformulate our approach to meet these requirements. 
Specifically, we propose a more general form of Eq.~\eqref{eq:volume_rendering}, which can be formalized as:
\begin{equation}
\label{eq:general_volume_rendering}
C(\mathbf{r})=\sum_{i=1}^{N}  \mathcal{W}\left(\Mat{F}_1, \dots,\Mat{F}_i\right) \mathcal{C}\left(\Mat{F}_i\right),
\end{equation}
where $\Mat{F}_i$ represents the set comprising image feature $f^{img}$ and volume feature $f^{v}$ in the $\Mat{x}_i \in \real^3$, $\mathcal{W}\left( \cdot \right)$ is the weight function that satisfies those three key properties we mentioned above, and and $\mathcal{C}\left( \cdot \right)$ denotes the color function. 
Specifically, color $c$ can then be interpreted as characteristic of each feature point in~\ref{eq:general_volume_rendering}. Therefore, feature at each point can be aggregated in a manner analogous to RGB value, and enabling us to deduce the primary feature points. This can be mathematically expressed as:
\begin{equation}\label{eq: deviate general vr}
    C(\mathbf{r})=C(\sum_{i=1}^{N}  \mathcal{W}\left(\Mat{F}_1, \dots,\Mat{F}_i\right)\Mat{F}_i),
\end{equation}
where the $C(\cdot)$ represents the color function that maps the feature into RGB space.

\subsection{Reconstruction Transformer}
\label{sec:trans_deocde}
Note that Eq.~\eqref{eq:volume_rendering} can be considered as a special form of  Eq. \eqref{eq:general_volume_rendering}. 
With this generalized form, we can reformulate the rendering function to overcome the oversimplification weakness of Eq.~\eqref{eq:volume_rendering}. 
Based on Eq. \eqref{eq:general_volume_rendering}, we introduce the Reconstruction Transformer (ReTR) that preserves essential rendering properties while incorporating sufficient complexity to implicitly learn and model intricate physical processes. 
ReTR is composed of a Render Transformer and an Occlusion Transformer. The Render Transformer leverages a learnable "meta-ray token" to encapsulate complex render properties, enhancing surface modeling. The Occlusion Transformer utilizes an attention mask to enable the occlusion-aware property. Also, ReTR works in the high-dimensional feature space instead of the color space, and thus allows for more complex and physically accurate light interactions. Consequently, ReTR not only overcomes the restrictions of Eq.~\eqref{eq:volume_rendering} but also preserves its fundamental rendering characteristics. We elaborate on the design of these components as follows. 

\mypara{Render Transformer.} Here, we discuss the design of the render transformer. Specifically, we introduce a global learnable token, refer to as "meta-ray token" and denote as $\mathbf{f}^{tok} \in \real^{D}$, to capture and store the complex render properties. For each sample ray, we first use the FeatureFusion block to combine all features associated with each sample point of the ray, resulting in $\mathbf{f}^{f}_i = \text{FeatureFusion}(\Mat{F}_i) \in \real^{D}$. We then employ the cross-attention mechanism within the Render Transformer to simulate the interaction of sample points along the ray. It can be formalized as follows:
\begin{equation}
    \label{eq:cross_attn_vr}
    C(\mathbf{r})=\mathcal{C}\left(\sum_{i=1}^{N} softmax\left( \frac{q(\mathbf{f}^{tok})k(\mathbf{f}^{f}_i)^\top}{\sqrt{D}}\right)v(\mathbf{f}^{f}_i)\right),
\end{equation}
where $q(\cdot),k(\cdot),v(\cdot)$ denotes three linear layers and $\mathcal{C}(\cdot)$ in this formulation is an MLP structure to directly regress observed color from the aggregated feature, and $W(\cdot)$ translates to $softmax\left( \frac{q(\mathbf{f}^{tok})k(\mathbf{f}^{f}_i)^\top}{\sqrt{D}}\right)$. Furthermore, $C(\cdot)$ is operationalized as MLP, which serves to decode the integrated feature into its corresponding RGB value. Then, the hitting probability will be normalized by $softmax$ function, which is \textit{Non-negative} and encourages the network to learn a weight distribution with positive kurtosis. And we can extract the attention map from the Render Transformer and derive the rendered depth as:
\begin{equation} \label{eqn:y}
D\left(\mathbf{r}\right) = \sum_{i=1}^{N} \alpha_{i}t_i,\\ \quad \text{where} \quad \alpha_{i} =  softmax\left( \frac{q(\mathbf{f}^{tok})k(\mathbf{f}^{f}_i)^\top}{\sqrt{D}}\right),
\end{equation}
where $\alpha_{1}, \dots , \alpha_{N}$ denotes the attention map extracted from the Eq.~\eqref{eq:cross_attn_vr}. The rendered depth map can be further used to generate mesh~\cite {curless1996volumetric} and point cloud~\cite{schonberger2016pixelwise}.

\mypara{Occlusion Transformer.} To further make our system \textit{Occlusion-aware} and enable of simulation of photon-medium interaction. We introduce Occlusion Transformer. Similar to previous works~\cite{wang2022attention,tang2023able}, we introduce an attention mask to achieve that the sample point interacts only with the points in front of it and the meta-ray token. Such unidirectional processes encourage the later points to respond to the preceding surface. This process can be formalized as:
\begin{equation} 
\label{eqn:RF_definition}
    \mathbf{R}^{f} = \{\mathbf{f}^{tok},\mathbf{f}^{f}_1,\mathbf{f}^{f}_2,\dots,\mathbf{f}^{f}_N\},
\end{equation}
\begin{equation} 
\label{eqn:occtrans}
\begin{aligned}
\mathbf{R}^{occ} = &\text{OccTrans}(Q,K,V = \mathbf{R}_f), \\ \quad \text{where} \quad \mathbf{f}_{i}^{occ} = \text{MLP}(\text{MHA}&(Q = \mathbf{f}^{f}_{i},K,V = \{\mathbf{f}^{tok},\mathbf{f}^{f}_{1}, \dots ,\mathbf{f}^{f}_{i}\})+\mathbf{f}^{f}_{i}).
\end{aligned}
\end{equation}
Where $\text{MHA}$ denotes the multi-head self-attention operation~\cite{vaswani2017attention} and $\mathbf{f}_{i}^{occ}$ denotes the refine feature of $\Mat{x}_i$ which obtained from the render transformer. In addition, $R^{f}$ is the collective set of tokens, and $R^{occ}$ signifies the occlusion transformer that employs $R^{f}$ for cross attention. Then, our Eq.~\eqref{eq:cross_attn_vr} can be rewritten as:
\begin{equation} 
\label{eqn:final_cross_attn_vr}
    C(\mathbf{r})=\mathcal{C}\left(\sum_{i=1}^{N} softmax\left( \frac{q(\mathbf{f}^{tok})k(\mathbf{f}_{i}^{occ})^\top}{\sqrt{D}}\right)v(\mathbf{f}_{i}^{f})\right).
\end{equation}

\mypara{Continuous Positional Encoding.} 
Following the traditional transformer design, we need to introduce a positional encoding to make the whole structure positional-aware. However, positional encoding proposed in~\cite{vaswani2017attention} ignores the \textit{actual distance} between each token, which is unsuitable for our situation. Furthermore, weighted-based resampling~\cite{mildenhall2021nerf} would lead to misalignment of positional encoding when an increased number of sample points are used.

To solve this problem, we extend the traditional positional encoding formula to continuous scenarios. Specifically, it can be formulated as follows:
\begin{equation}
\begin{aligned}
    PE_{(\Mat{x}_i,2i)} = sin(\beta t_i / 10000^{2i/D}), \\
    PE_{(\Mat{x}_i,2i+1)} = cos(\beta t_i / 10000^{2i/D}).
\end{aligned}
\end{equation}
Here, $i$ represents the positional encoding in the $i_{th}$ dimension and $\beta$ is a scale hyperparameter we empirically set to 100. The updated formula successfully solves the misalignment of traditional positional encoding, results are shown in Tab.~\ref{table:ablation_sampling}. The specific proofs will be included in the Appendix section.

\begin{figure*}[!t]
\begin{center}
\setlength{\abovecaptionskip}{0.cm}
\centerline{\includegraphics[width=1.\textwidth]{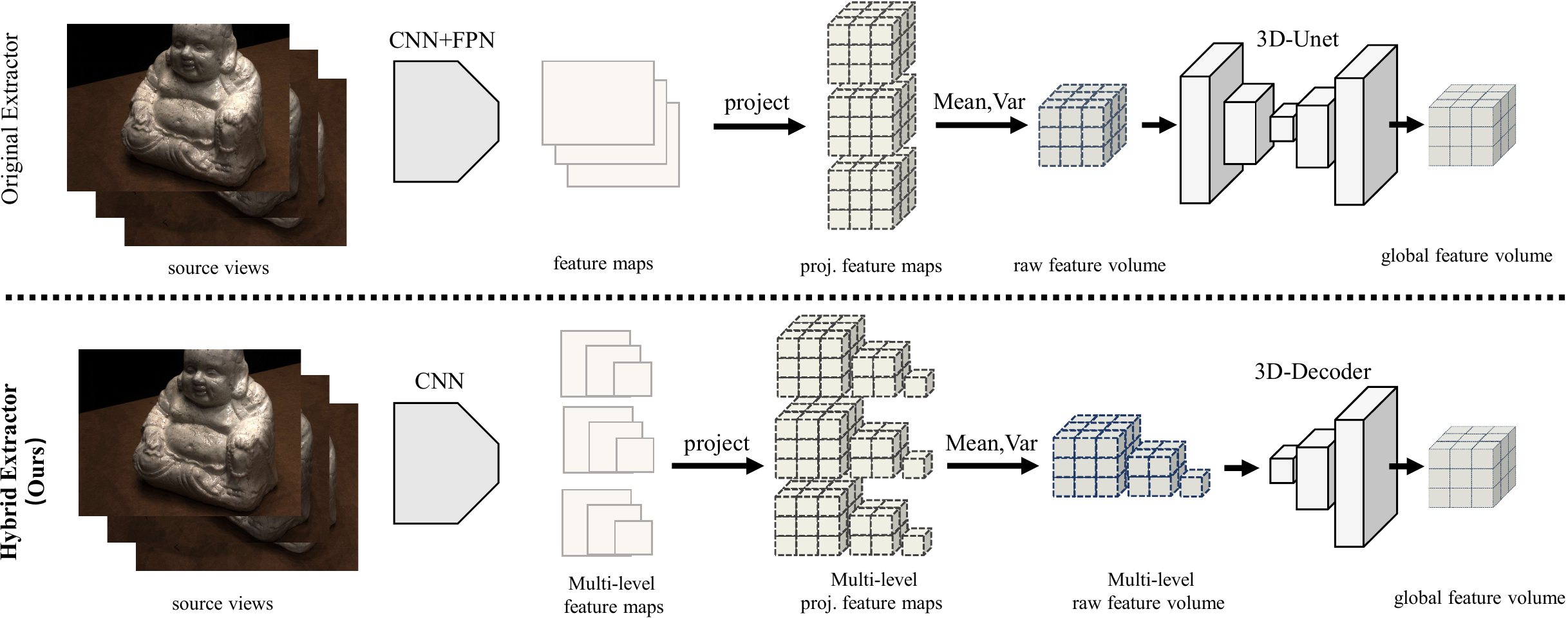}}
\vspace{0.3cm}
\caption{Comparision of the original extractor used in~\cite{ren2022volrecon,long2022sparseneus} (top) and our \textbf{hybrid extractor}~(bottom). The original extractor primarily discerns high-level features, demonstrating less efficacy. In contrast, our hybrid extractor excels in integrating multi-level features, demonstrating superior efficacy.}
\label{fig:he}
\end{center}
\vspace{-0.5cm}
\end{figure*}
\subsection{Hybrid Extractor}
\label{sec.hbe}
In our learning-based rendering, a finer level of visual feature is necessary, which is not achieved by traditional feature extractors that rely on high-level features obtained through FPN. These features are highly semantic abstract and not suitable for low-level visual feature matching~\cite{cao2022mvsformer}. To overcome this limitation, inspired by NeuralRecon~\cite{sun2021neuralrecon}, we further propose Hybrid Extractor. Rather than relying on FPN to generate one feature map from high-level features of CNN, and using a 3D U-Net to process projected features as shown in Fig.~\ref{fig:he}, we leverage all level features from various layers to construct multi-level volume features. Then, we adopt a 3D CNN decoder to fuse and decode the multi-level volume features, producing the final global volume features. 

Our approach enables us to perceive both low and high-level features, which is crucial for generalizable neural surface reconstructions that require detailed surface processing. Second, by avoiding the use of the encoder part of the 3D U-Net, we reduce the computational complexity and allow us to build a higher resolution volume feature within the same computational budget. 

\subsection{Loss Functions}
\label{sec:netarch}
Our overall loss function is defined as the weighted sum of two loss terms:
\begin{equation}
\label{total_loss}
\mathcal{L}= \mathcal{L}_{\text {rendering}}+\alpha \mathcal{L}_{\text {depth}},
%+ \gamma \mathcal{L}_{\text {box}}.
\end{equation}
where $\mathcal{L}_{\text {rendering}}$ constrains the observed colors to match the ground truth colors and is formulated as:
\begin{equation}
    \mathcal{L}_{\text {rendering}} = \frac{1}{S} \sum_{s=1}^S \left\|C\left( \mathbf{r} \right)-C_{g}\left( \mathbf{r} \right)\right\|_2,
\end{equation}
Here, $S$ is the number of sampled rays for training, and $C_{g}\left( \mathbf{r} \right)$ represents the ground truth color of the sample ray $r$. The depth loss $\mathcal{L}_{\text{depth}}$ is defined as
\begin{equation}
    \mathcal{L}_{\text {depth}} = \frac{1}{S_1} \sum_{s=1}^{S_1} |D\left( \mathbf{r} \right)-D_{g}\left( \mathbf{r} \right)|,
\end{equation}
where $S_1$ is the number of pixels with valid depth and $D_{g}\left( \mathbf{r} \right)$ is the ground truth depth. 
In our experiments, we set $\alpha=1.0$. 

%%%%%%%%%%%%%%%%%   EXP    %%%%%%%%%%%%%%%%%%%%%%%%%%
%%%fig and tab
\section{Experiments}
\mypara{Datasets.} The DTU dataset~\cite{aanaes2016large} is a large-scale indoor multi-view stereo dataset consisting of 124 different scenes captured under 7 different lighting conditions. To train our frameworks, we adopted the same approach as in previous works~\cite{long2022sparseneus, ren2022volrecon}. Furthermore, we evaluated our models' generalization capabilities by testing them on three additional datasets: Tanks \& Templates~\cite{Knapitsch2017}, ETH3D~\cite{schops2017multi}, and BlendedMVS~\cite{yao2020blendedmvs}, where no additional training was performed on the testing datasets.

\begin{figure*}[!t]
\vspace{-2em}
% \vskip 0.2in
\begin{center}
\setlength{\abovecaptionskip}{0.cm}
\centerline{\includegraphics[width=1.\textwidth]{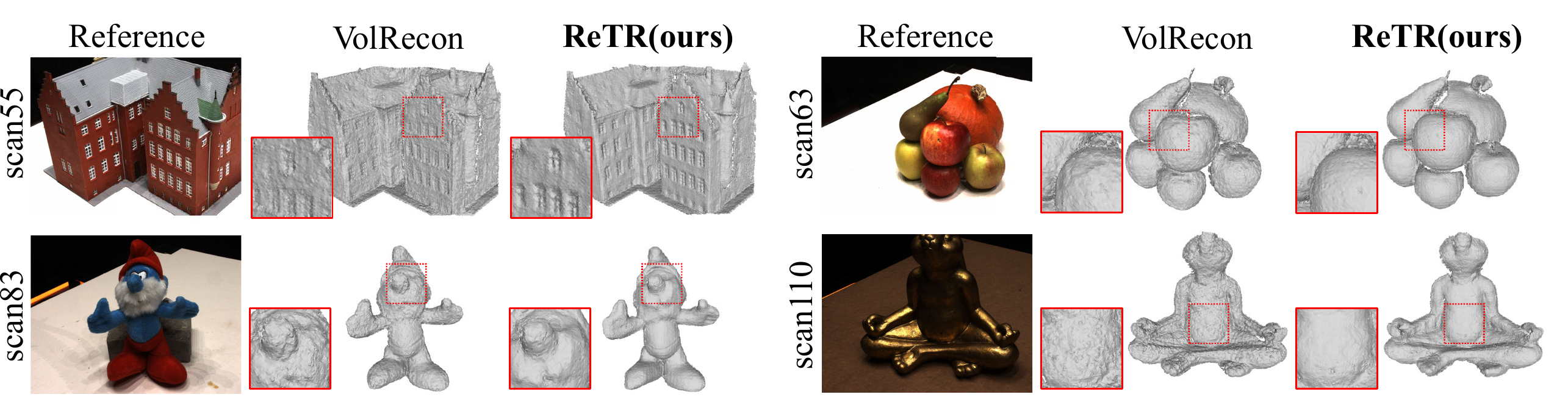}}
\caption{Sparse view reconstruction on testing scenes in the DTU~\cite{aanaes2016large}. Comparison with VolRecon~\cite{ren2022volrecon} (left), our proposed ReTR (right) renders a more accurate surface and preserves finer details, \emph{e.g.} the window of the house (scan 24), the nose of smurf, ReTR produces much sharper details. Best viewed on a screen when zoomed in.}
\vspace{-0.5cm}
\label{fig:sparserecon}
\end{center}
\end{figure*}

\begin{table*}[!t]
        \vspace{-0.2cm}
        \resizebox{\textwidth}{!}{%
        \renewcommand{\arraystretch}{1.1}
        \centering
        \small
        \begin{tabular}{lcccccccccccccccc} \\  \toprule
SCAN         & \multicolumn{1}{l}{Mean$ \downarrow$} & 24   & 37   & 40   & 55   & 63   & 65   & 69   & 83   & 97   & 105  & 106  & 110  & 114  & 118  & 122  \\ \hline
COLMAP~\cite{schonberger2016pixelwise} & 1.52 & \textbf{0.90} & 2.89 & 1.63 & 1.08 & 2.18 & 1.94 & 1.61 & \underline{1.30} & 2.34 & 1.28 & 1.10 & 1.42 & 0.76 & 1.17 & \underline{1.14} \\ 
MVSNet~\cite{yao2018mvsnet} & \underline{1.22} & \underline{1.05} & \underline{2.52} & 1.71 & 1.04 & 1.45 & \textbf{1.52} & \textbf{0.88} & \textbf{1.29} & \underline{1.38} & \underline{1.05} & \textbf{0.91} & \textbf{0.66} & 0.61 & \underline{1.08} & 1.16 \\
\hline
IDR \cite{yariv2020multiview} & 3.39 & 4.01 & 6.40 & 3.52 & 1.91 & 3.96 & 2.36 & 4.85 & 1.62 & 6.37 & 5.97 & 1.23 & 4.73 & 0.91 & 1.72 & 1.26 \\
VolSDF \cite{yariv2021volume} & 3.41 & 4.03 & 4.21 & 6.12 & \textbf{0.91} & 8.24 & \underline{1.73} & 2.74 & 1.82 & 5.14 & 3.09 & 2.08 & 4.81 & \underline{0.60} & 3.51 & 2.18 \\
UNISURF \cite{oechsle2021unisurf} & 4.39 & 5.08 & 7.18 & 3.96 & 5.30 & 4.61 & 2.24 & 3.94 & 3.14 & 5.63 & 3.40 & 5.09 & 6.38 & 2.98 & 4.05 & 2.81 \\
NeuS \cite{wang2021neus} & 4.00 & 4.57 & 4.49 & 3.97 & 4.32 & 4.63 & 1.95 & 4.68 & 3.83 & 4.15 & 2.50 & 1.52 & 6.47 & 1.26 & 5.57 & 6.11 \\ \hline

PixelNeRF \cite{yu2021pixelnerf}   & 6.18                     & 5.13 & 8.07 & 5.85 & 4.40  & 7.11 & 4.64 & 5.68 & 6.76 & 9.05 & 6.11 & 3.95 & 5.92 & 6.26 & 6.89 & 6.93 \\
IBRNet \cite{wang2021ibrnet}      & 2.32                     & 2.29 & 3.70  & 2.66 & 1.83 & 3.02 & 2.83 & 1.77 & 2.28 & 2.73 & 1.96 & 1.87 & 2.13 & 1.58 & 2.05 & 2.09 \\
MVSNeRF \cite{chen2021mvsnerf}     & 2.09                     & 1.96 & 3.27 & 2.54 & 1.93 & 2.57 & 2.71 & 1.82 & 1.72 & 2.29 & 1.75 & 1.72 & 1.47 & 1.29 & 2.09 & 2.26 \\ \hline

SparseNeuS \cite{long2022sparseneus}   & 1.96                     & 2.17 & 3.29 & 2.74 & 1.67 & 2.69 & 2.42 & 1.58 & 1.86 & 1.94 & 1.35 & 1.50  & 1.45 & 0.98 & 1.86 & 1.87 \\
VolRecon \cite{ren2022volrecon}& 1.38   & 1.20	& 2.59 &	\underline{1.56}	& 1.08 &	\underline{1.43} &	1.92 &	\underline{1.11} & 1.48 & 1.42 & \underline{1.05} & 1.19 & 1.38 & 0.74 & 1.23 & 1.27  \\ 
\textbf{ReTR (Ours)} & \textbf{1.17}   & \underline{1.05}	& \textbf{2.31} &	\textbf{1.44}	& \underline{0.98} &	\textbf{1.18} &	\textbf{1.52} &	\textbf{0.88} & 1.35 & \textbf{1.30} & \textbf{0.87} & \underline{1.07} & \underline{0.77} & \textbf{0.59} & \textbf{1.05} & \textbf{1.12}  \\
\bottomrule
\end{tabular}

% \textbf{ReTR (base)} & \textbf{1.16}   & \underline{0.98}	& \textbf{2.27} &	\textbf{1.59}	& \underline{1.00} &	\textbf{1.14} &	\textbf{1.56} &	\textbf{0.90} & 1.35 & \textbf{1.26} & \textbf{0.86} & \underline{1.06} & \underline{0.78} & \textbf{0.57} & \textbf{1.01} & \textbf{1.07}  \\ 
% \textbf{ReTR (large)} & \textbf{1.15}   & \underline{0.96}	& \textbf{2.27} &	\textbf{1.64}	& \underline{0.95} &	\textbf{1.19} &	\textbf{1.59} &	\textbf{0.86} & 1.32 & \textbf{1.25} & \textbf{0.85} & \underline{1.02} & \underline{0.75} & \textbf{0.55} & \textbf{1.02} & \textbf{1.11}  \\ 
}
        \caption{Quantitative results of \textbf{sparse view} reconstruction on 15 testing scenes of DTU dataset~\cite{aanaes2016large}. We report the chamfer distance, the lower the better, Methods are split into four categories from top to bottom: a) MVS-based methods, b) Per-scene optimization methods, c) Generalizable rendering methods, and d) Generalizable reconstruction methods. The best scores are in \textbf{bold} and the second best are in \underline{underlined}.}
        \label{table::fewviewchamfer}
        \vspace{-0.4cm} 
\end{table*}

\mypara{Baselines.} To demonstrate the effectiveness of our method from various perspectives, we compared it with (1) SparseNeus~\cite{long2022sparseneus} and VolRecon~\cite{ren2022volrecon}, the state-of-the-art generalizable neural surface reconstruction method; (2) Generalizable neural rendering methods (3) Neural implicit reconstruction~\cite{yariv2020multiview,wang2021neus,yariv2021volume,oechsle2021unisurf} which require individual training for each scene from scratch. (4) Popular multi-view stereo (MVS)~\cite{schonberger2016pixelwise,yao2018mvsnet} methods. Further details on the baselines are provided in the appendix.
\subsection{Sparse View Reconstruction}
For comparison, we performed sparse reconstruction using only three views, following the same approach as~\cite{ren2022volrecon,long2022sparseneus}. We adopted the same evaluation process and testing split as used in previous works~\cite{ren2022volrecon,long2022sparseneus} to ensure a fair comparison. We use a similar approach as VolRecon~\cite{ren2022volrecon} to generate mesh, more details can be found in Appendix.
As shown in Tab.~\ref{table::fewviewchamfer}, our method outperforms VolRecon~\cite{ren2022volrecon} and SparseNeuS~\cite{long2022sparseneus} by a significant margin. Moreover, our method also outperforms popular MVS methods such as MVSNet~\cite{yao2018mvsnet}. Furthermore, we present the qualitative results of sparse view reconstruction in Fig.~\ref{fig:sparserecon}. Our reconstructed geometry exhibits smoother surfaces and less noise compared to the current SoTA methods.

\subsection{Depth Map Evaluation \& Full View Reconstruction} 
\begin{figure*}[t]
%\vskip 0.2in
\begin{center}
\setlength{\abovecaptionskip}{0.cm}
\centerline{\includegraphics[width=1.\textwidth]{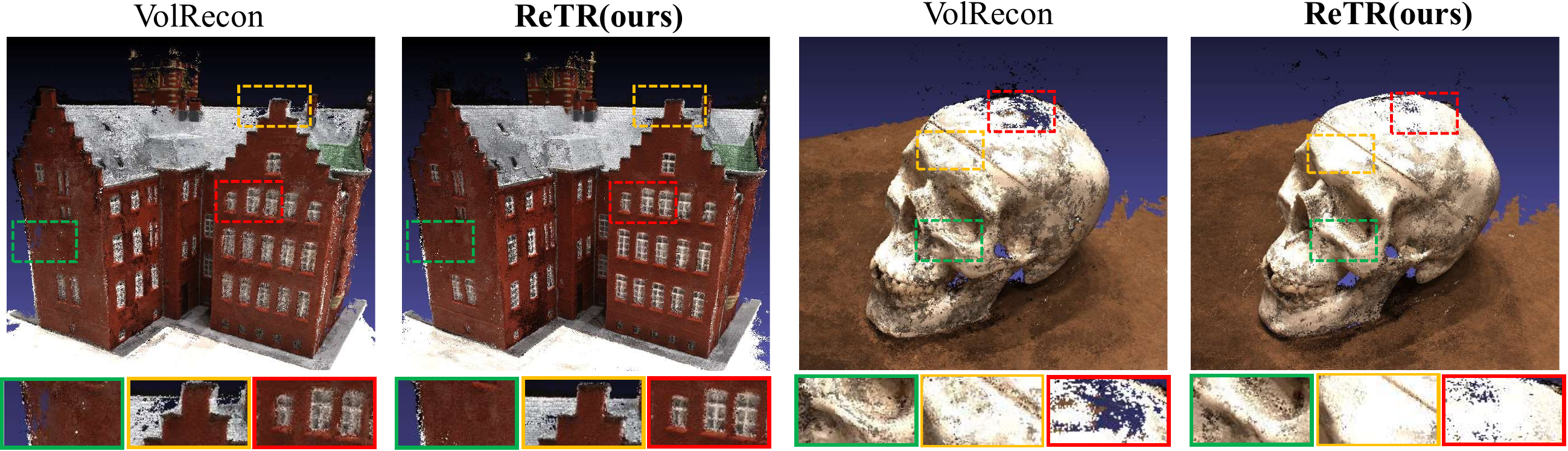}}
\caption{Full view reconstruction visualization in test set of DTU~\cite{aanaes2016large}, comparison with VolRecon~\cite{ren2022volrecon} (left), our proposed ReTR (right) reconstructs better point clouds, \emph{e.g.} fewer holes, the skull head top, and the house roof gives a much complete representation, \emph{e.g.} finer details, the house window, and the skull cheek, provides much finer details. Best viewed on a screen when zoomed in.}
\vspace{-6mm}
\label{fig:fullrecon}
\end{center}
\end{figure*}

We compare our rendered depth with those generated by SparseNeuS~\cite{long2022sparseneus}, MVSNet~\cite{yao2018mvsnet} and VolRecon~\cite{ren2022volrecon}. Following the experiment settings introduced in VolRecon~\cite{ren2022volrecon}, we also use four source views as input for depth rendering. Additionally, we evaluated the performance by fusing all depth maps into a global point cloud. As shown in Tab.~\ref{table:pc_depth_table}, our method outperforms existing methods in both evaluations. Moreover, as demonstrated in Fig.~\ref{fig:fullrecon}, our method achieves a sharper boundary with less noise and fewer holes compared to the current SoTA method~\cite{ren2022volrecon}.

\begin{table*}[t]
        %\resizebox{\textwidth}{!}{%
        \begin{minipage}{1.\textwidth}
        \renewcommand{\arraystretch}{1.1}
        \centering
        \scriptsize
        \begin{tabular}{M{16mm} M{14mm} M{14mm} M{14mm} |M{12mm} M{12mm} M{12mm}}
\toprule
Method & {Acc. $\downarrow$} & {Comp. $\downarrow$} & {Chamfer $\downarrow$} & {$<$1mm $\uparrow$} & {$<$2mm $\uparrow$} & {$<$4mm $\uparrow$} \\ \midrule
MVSNet~\cite{yao2018mvsnet} & 0.55 & 0.59 & 0.57 & 29.95 & 52.82 & 72.33 \\
SparseNeuS~\cite{long2022sparseneus} & 0.75 & 0.76 & 0.76 & 38.60 & 56.28 & 68.63 \\
VolRecon~\cite{ren2022volrecon} & 0.55 & 0.66 & 0.60 & 44.22 & 65.62 & 80.19 \\
\textbf{ReTR (Ours)}   & \textbf{0.54}     & \textbf{0.51}      & \textbf{0.52}        & \textbf{45.00} & \textbf{66.43} & \textbf{81.52} \\
\bottomrule
\end{tabular}%%}
        % \vspace{-0.1cm}
        \caption{Quantitative results of \textbf{full view} reconstruction on 15 testing scenes of DTU dataset~\cite{aanaes2016large}. For the Accuracy (ACC), Completeness (COMP), and Chamfer Distance, the lower is the better. For depth map evaluation, threshold percentages (<1mm, <2mm, <4mm) are reported in percentage (\%). The best scores are in \textbf{bold}.}
        \label{table:pc_depth_table}
        \end{minipage}
        \vspace{-0.5cm}
\end{table*}

\subsection{Generalization}
\begin{figure*}[t]
%\vspace{-2em}
\begin{center}
\setlength{\abovecaptionskip}{0.cm}
\centerline{\includegraphics[width=1.\textwidth]{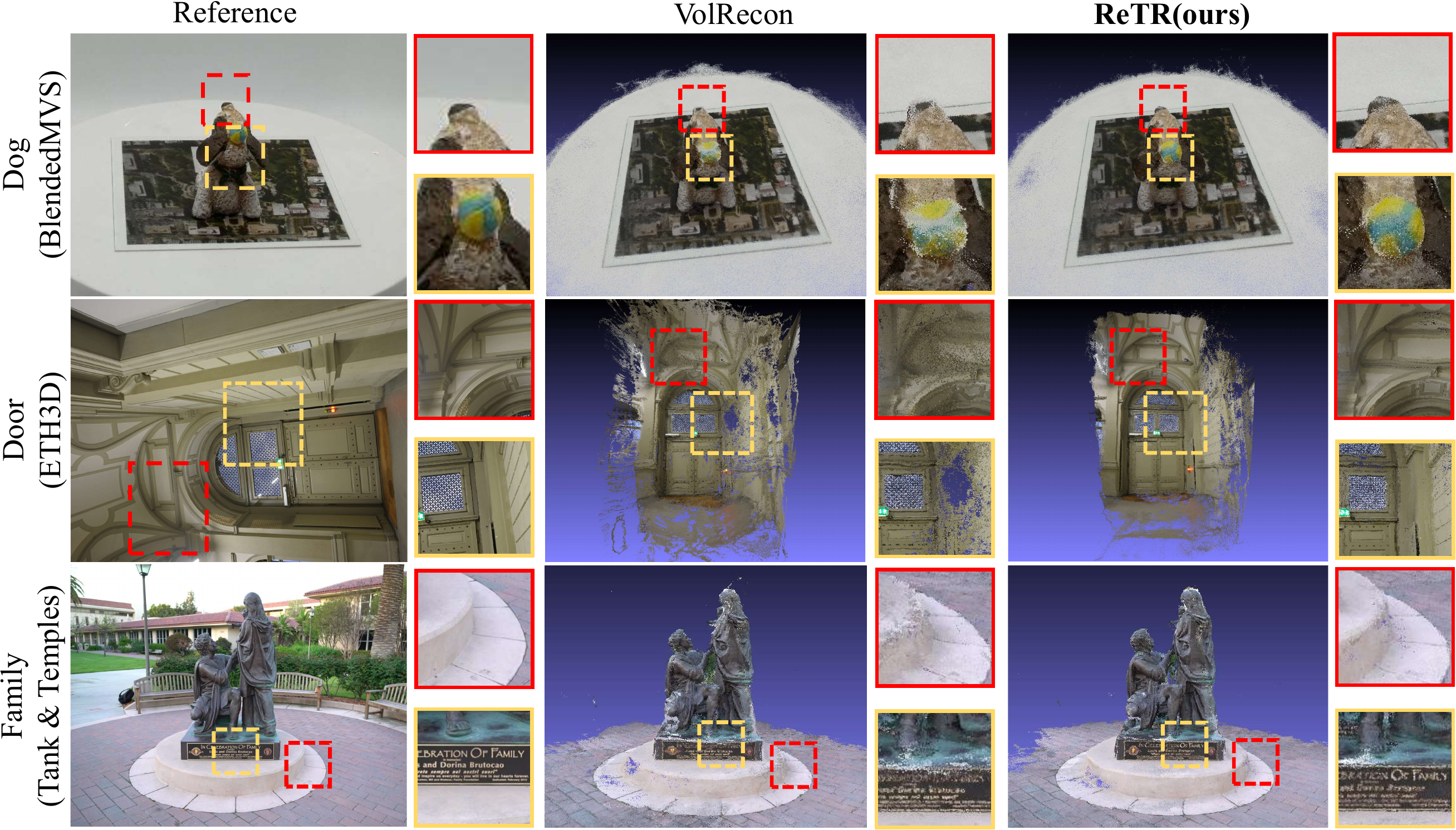}}
\vspace{0.1cm}  
\caption{Generalize reconstruction visualization in 3 datasets BlendedMVS, ETH3D, and Tank \& Temples~\cite{yao2020blendedmvs,aanaes2016large,schops2017multi}, comparison with VolRecon~\cite{ren2022volrecon} (middle), our proposed ReTR (right) generalize well to the large-scale datasets without fine-tuning. ReTR produces finer details and sharper boundaries. Best viewed on a screen when zoomed in.}
\label{fig:grecon}
\vspace{-0.5cm}
\end{center}
\end{figure*}

To evaluate the generalization ability of our model without retraining, we use three datasets, namely Tank \& Temples, BlendedMVS, and ETH3D~\cite{Knapitsch2017, yao2020blendedmvs, schops2017multi}. The high-quality reconstruction of large-scale scenes and small objects in different domains, as shown in Fig.~\ref{fig:grecon}, demonstrates the effectiveness of our method in terms of generalization capability.
\section{Ablation Study}
\label{sec.ablation}
We conduct ablation studies to examine the effectiveness of each module in our design. The ablation study results are reported on sparse view reconstruction in test split following SparseNeuS~\cite{long2022sparseneus} and VolRecon~\cite{ren2022volrecon}.

\begin{wraptable}{h}{0.40\linewidth}
        \vspace{-0.2cm}
        \centering
        \scriptsize
        \begin{tabular}{ M{7mm} M{7mm} M{7mm} M{12mm} }
\toprule
Reder Trans. & Occ. Trans. & Hybrid Ext. & Chamfer$\downarrow$ \\
\specialrule{0em}{0pt}{1pt}
\hline
\specialrule{0em}{0pt}{1pt}
% $\times$ &  $\times$ & $\times$ & 1.42\\
$\checkmark$ & $\times$ &  $\times$& 1.31 \\
$\checkmark$ & $\times$ & $\checkmark$ & 1.29\\
$\checkmark$ & $\checkmark$ & $\times$ & 1.28\\
$\checkmark$ & $\checkmark$ & $\checkmark$ & \textbf{1.17}\\
\bottomrule
\end{tabular}

% \begin{tabular}{ M{4mm} M{8mm} M{8mm} M{8mm} M{10mm} M{8mm} }
%     \toprule
%     ID      & GVA        & PE Mul     & Grid Pool & Map Unpool  & mIoU (\%)     \\
%     \specialrule{0em}{0pt}{1pt}
%     \hline
%     \specialrule{0em}{0pt}{1pt}
%     \Rom{1} &            &            &            &            & 72.3          \\
%     \Rom{2} & \Checkmark &            &            &            & 73.8          \\
%     \Rom{3} & \Checkmark & \Checkmark &            &            & 74.4          \\
%     \Rom{4} & \Checkmark & \Checkmark & \Checkmark &            & 74.9          \\
%     \Rom{5} & \Checkmark & \Checkmark & \Checkmark & \Checkmark & \textbf{75.4} \\
%     \bottomrule
% \end{tabular}

        \caption{Model component ablation. All of these parts are described in Sec.~\ref{sec:trans_deocde}.}
        \label{tab:ablation_part}
        \vspace{-0.3cm}
\end{wraptable}

\mypara{Effectiveness of Modules.} We evaluate key components of our approach to generalizable neural surface reconstruction, as shown in Tab.~\ref{tab:ablation_part}. 
For the evaluation of the occlusion transformer, we keep the original transformer architecture while removing the special design we proposed in Sec.~\ref{sec:trans_deocde}, to ensure the training parameter would not affect the evaluation. For the hybrid extractor part, we replace this module with the original extractor that has been used in~\cite{ren2022volrecon}. Our results demonstrate that our approach can better aggregate features from different levels and use them more effectively. These evaluations highlight the importance of these components in our approach.

\mypara{Robustness of Different Sampling.} 
Tab.~\ref{table:ablation_sampling} displays the effects of altering the number of sample points on the reconstruction quality of VolRecon\cite{ren2022volrecon} and ReTR. Our method surpasses the current SoTA, even when the number of sampling points decreases. These results suggest that existing methods that rely on sampling points of the ray struggle to provide confident predictions of the surface due to the nature of volume rendering. Our approach, which uses learning-based rendering, is more resilient to sampling strategies and can provide reliable depth estimations even with fewer samples. Meanwhile, the effectiveness of continuous P.E. in Sec.~\ref{sec:trans_deocde} is proved through the result.

\mypara{Unsupervised  Neural Surface Reconstruction.} Our approach is still applicable to unsupervised neural surface reconstruction using only colors for training, which remove $ \mathcal{L}_{\text {depth}}$. Meanwhile, we find that our method significantly outperforms the current SoTA method under unsupervised situations and is even comparable to COLMAP~\cite{schonberger2016pixelwise} the popular MVS technique, As shown in Tab.~\ref{tab:ablation_unsup}, This is further evidence that the improvement of complex rendering systems for implicit reconstruction is huge.

\section{Conclusion}
We have proposed Reconstruction Transformer (ReTR), a novel framework for generalizable neural surface reconstruction that uses transformers to model complex rendering processes. ReTR represents a significant advancement in the field of surface reconstruction, offering a powerful solution to the challenges faced by neural implicit reconstruction methods. Additionally, we delve into the design procedure of learning-based rendering. This exploration broadens our understanding of enhancing complex rendering systems and sets the stage for future research endeavors not only in surface reconstruction but also in other tasks relative to differentiable rendering.

\section{Acknowledgements}
We thank Shuai Yang and Wenhang Ge for the thoughtful review of our manuscript and valuable discussions throughout this project. Thank you to Yukang Chen, Shuhan Zhong and Jierun Chen for ideas and feedbacks on our manuscript. We would also like to thank the Turing AI Computing Cloud (TACC)~\cite{tacc} and HKUST iSING Lab for providing us with computation resources on their platform.
\begin{table}[t] 
    \begin{minipage}{.47\textwidth}
        \centering
        \scriptsize
        % \begin{tabular}{cccc}
% \hline
% \multicolumn{1}{c}{\multirow{2}{*}{Method}} & \multicolumn{2}{c}{Chamfer} & \multicolumn{1}{c}{\multirow{2}{*}{\Delta \%}} \\
% \multicolumn{1}{c}{} & 64+64 & 64+0  & \multicolumn{1}{c}{} \\ \hline
% VolRecon~\cite{ren2022volrecon}  & 1.38 & 1.45 & 4.83                \\
% Ours                 & 1.18 & 1.21 & \textbf{2.48}              \\ \hline
% \end{tabular}%

% \begin{tabular}{ M{12mm} M{12mm} M{12mm} M{12mm} }
% \hline
% \multirow{2}{*}{\begin{tabular}[c]{@{}c@{}} Points\end{tabular}} & \multicolumn{2}{c}{Chamfer $\downarrow$} & \multicolumn{1}{c}{\multirow{2}{*}{$\Delta$ \%}} \\
%   & VolRecon~\cite{ren2022volrecon} & Ours          & \multicolumn{1}{c}{}    \\ \hline
% 16+16 & 1.60     & 1.35          & -15.6                   \\
% 32+32 & 1.45     & 1.21          & -16.6                   \\
% 64+0  & 1.45     & 1.26          & -13.1                   \\
% 64+64 & 1.38     & 1.18          & -14.5                    \\ 
% 128+0 & 1.79     & 1.26          & -29.6                    \\\hline
% \end{tabular}%

% \begin{tabular}{ M{12mm} M{15mm} M{15mm} M{4mm} }
% \toprule
% Sample Points &VolRecon\cite{ren2022volrecon} & \textbf{Ours} & $\Delta$ $\%$ \\
% \specialrule{0em}{0pt}{1pt}
% \hline
% \specialrule{0em}{0pt}{1pt}
%  [$16,16$] & 1.72     & 1.36          & -20.9             \\
%  [$32,32$] & 1.38     & 1.18          & -14.5              \\
%  [64,64] & 1.35     & 1.13          & -16.3               \\
%  [] & 1.33     & \textbf{1.09} & \multicolumn{1}{c}{-18} \\ \bottomrule
% \end{tabular}%
\begin{tabular}{M{12mm} M{14mm} M{14mm} M{8mm} }
\toprule
Sample Points & VolRecon \cite{ren2022volrecon} & {ReTR (Ours$^*$)} & \textbf{ReTR (Ours)} \\
\specialrule{0em}{0pt}{1pt}
\hline
\specialrule{0em}{0pt}{1pt}
 $(16,16)$ & 1.60     & 1.59          & 1.30              \\
 $(32,32)$ & 1.45     & 1.42          & 1.20              \\
 $(64,0)$ & 1.45      & 1.26           & 1.22               \\
 $(128,0)$ & 1.79     & 1.46          & 1.20                   \\ 
 $(64,64)$ & 1.38     & 1.18     & \textbf{1.17}           \\
\bottomrule
\end{tabular}%
        \caption{Number of sampling ablation. The $*$ denotes our model using traditional positional encoding.}
        \label{table:ablation_sampling}
    \end{minipage}
    \begin{minipage}{.52\textwidth}
        \centering
        \vspace{-3mm}
        \scriptsize
        % \begin{tabular}{ccc}
% \hline
% % \multicolumn{1}{c}{\multirow{2}{*}{Method}} & \multicolumn{2}{c}{Mean Chamfer $\downarrow$} & \multicolumn{1}{c}{\multirow{2}{*}{\Delta \%}} \\
% \multicolumn{1}{c}{Method} & w$\mathcal{L}_{\text {depth}}$ & w/o $\mathcal{L}_{\text {depth}}$\\ \hline
% SparseNeuS~\cite{long2022sparseneus}   & 4.22 & 1.96              \\
% VolRecon~\cite{ren2022volrecon}     & 1.38 & 2.06            \\ 
% Ours                 & 1.18 & 1.45       \\ \hline
% \end{tabular}%
% \begin{tabular}{M{18mm} M{24mm} M{14mm}}
%     \toprule
%     Local Type                             & Training Losses & Chamfer $\downarrow$      \\
%     \midrule
\begin{tabular}{M{16mm} M{22mm} M{14mm}}
    \toprule
    Method                            & Training Losses & Chamfer $\downarrow$      \\
    \midrule
    \multirow{2}{*}{SparseNeuS~\cite{long2022sparseneus}}&w$\mathcal{L}_{\text {depth}}$  & 4.22          \\
                                                       &w/o $\mathcal{L}_{\text {depth}}$ & 1.96          \\
    \midrule
    \multirow{2}{*}{VolRecon~\cite{ren2022volrecon}} &w$\mathcal{L}_{\text {depth}}$ & 1.38  \\
                                                    &w/o $\mathcal{L}_{\text {depth}}$ & 2.06\\
    \midrule
    \multirow{2}{*}{\textbf{ReTR (Ours)}} &w$\mathcal{L}_{\text {depth}}$& \textbf{1.17} \\
                                           &w/o $\mathcal{L}_{\text {depth}}$ & 1.45 \\
    \bottomrule
\end{tabular} \\
        \caption{Ablation study of training losses.}
        \label{tab:ablation_unsup}
    \end{minipage} \\
    \vspace{-7mm}    
\end{table}

\clearpage
% {\small
% \bibliographystyle{ieee_fullname}
% \bibliography{egbib}
% }

\bibliography{egbib}

%\iffalse
% \clearpage
% \appendix
% \input{appendix}

% \clearpage

% \input{rebuttal}

\end{document}